\documentclass[letterpaper, 10 pt, journal, twoside]{IEEEtran} 
\usepackage{amsmath,amssymb,amsopn,amstext,amsfonts} 
\usepackage{lmodern}
\usepackage{times}
\usepackage{multirow}
\usepackage{graphicx}
\usepackage{cancel}
\usepackage{balance}
\usepackage{xcolor}
\usepackage{mathtools} 
\usepackage{algorithm}
\usepackage{algorithmic}
\usepackage{bm}

\usepackage{diagbox}
\usepackage{float}
\usepackage{url}
\usepackage{booktabs}
\usepackage{framed}
\usepackage{stfloats}
\usepackage{cancel}
\usepackage[caption=false,font=normalsize,labelfont=sf,textfont=sf]{subfig}
\usepackage[normalem]{ulem}

\usepackage{gensymb}
\usepackage{siunitx}
\usepackage{microtype}
\usepackage{hyperref}
\usepackage{verbatim}

\graphicspath{{./Figures/}}
\DeclareGraphicsExtensions{.pdf,.png,.jpg,.eps,.svg}

\IEEEoverridecommandlockouts

\title{
FlyAware: Inertia-Aware Aerial Manipulation via Vision-Based Estimation and Post-Grasp Adaptation
}

\author{
Biyu Ye$^{1}$, Na Fan$^{2}$, Zhengping Fan$^{1}$, Weiliang Deng$^{1}$, Hongming Chen$^{1}$, Qifeng Chen$^{2}$, and Ximin Lyu$^{1,3}$%

\thanks{Manuscript received: June 27, 2025; Revised December 15, 2025; Accepted January 21, 2026. This paper was recommended for publication by Editor Soon-Jo Chung upon evaluation of the Associate Editor and Reviewers’comments. This work is supported by the National Key Research and Development Program of China (Grant No. 2023YFB4706600), the National Natural Science Foundation of China (Grant No.62303495), the Research Grants Council of HKSAR (Grant No. AoE/E-601/24-N), and the Young Talent Support Project of Guangzhou Association for Science and Technology (Grant No. QT-2025-004). \textit{(Biyu Ye and Na Fan are co-first authors.) (Corresponding author: Ximin Lyu.)}
(Project page: \url{https://flyaware.github.io/})}

\thanks{$^{1}$Biyu Ye, Zhengping Fan, Weiliang Deng, Hongming Chen and Ximin Lyu are with the School of Intelligent Systems Engineering, Sun Yat-sen University, Guangzhou 510275, China (e-mail: yeby9@mail2.sysu.edu.cn; fanzhp@mail.sysu.edu.cn; dengwliang@mail2.sysu.edu.cn; chenhm223@mail2.sysu.edu.cn; lvxm6@mail.sysu.edu.cn).}

\thanks{$^{2}$Na Fan and Qifeng Chen are with Visual Intelligence Lab, Cheng Kar-Shun Robotics Institute, The Hong Kong University of Science and Technology, Hong Kong, SAR, China (e-mail: nfanaa@cse.ust.hk; cqf@cse.ust.hk).}

\thanks{$^{3}$Ximin Lyu is also with Differential Robotics Technology Company, Ltd., Hangzhou 311121, China.}

\thanks{Digital Object Identifier: see top of this page.}
}

\begin{document}
\maketitle

\begin{abstract}
Aerial manipulators (AMs) are gaining increasing attention in automated transportation and emergency services due to their superior dexterity compared to conventional multirotor drones. However, their practical deployment is challenged by the complexity of time-varying inertial parameters, which are highly sensitive to payload variations and manipulator configurations. Inspired by human strategies for interacting with unknown objects, this letter presents a novel onboard framework for robust aerial manipulation. The proposed system integrates a vision-based pre-grasp inertia estimation module with a post-grasp adaptation mechanism, enabling real-time estimation and adaptation of inertial dynamics. For control, we develop an inertia-aware adaptive control strategy based on gain scheduling, and assess its robustness via frequency-domain system identification. Our study provides new insights into post-grasp control for AMs, and real-world experiments validate the effectiveness and feasibility of the proposed framework.
\end{abstract}


\section{INTRODUCTION}
\label{INTRODUCTION}
\begin{figure}[t]
\centering
\includegraphics[width=.99\columnwidth]{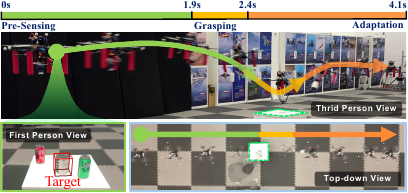}
\caption{Real-world experiment: The aerial manipulator pre-senses and grasps objects arbitrarily placed on a table by a human operator.}
\label{fig:one}
\vspace{-0.8cm}
\end{figure}

Aerial manipulators integrate the agility of drones with the dexterity of robotic arms, enabling complex tasks like grasping (Fig.~\ref{fig:one}) and retrieving objects in hard-to-reach environments—such as cliffs, rooftops, or under bridges~\cite{2018RAL_Review_Ruggiero, 2022TRO_PPFARM_Ollero}. Despite their potential, achieving precise and robust control remains a significant challenge. In particular, real-time estimation of geometric and inertial parameters—including mass, center of mass (CoM), and moment of inertia (MoI)—is crucial for stable flight: mass governs translational behavior, CoM affects stability, and MoI dictates attitude dynamics~\cite{2019ICRA_Online_Wuest}. These parameters, however, can vary significantly due to payload changes or manipulator reconfiguration, leading to estimation delays, prior knowledge gaps, and insufficient controller responsiveness during aerial interactions. Consequently, fast and accurate estimation becomes essential for maintaining reliable performance in dynamic, uncertain environments.

To cope with the unknown payload, 
model-based approaches use force/torque or IMU sensors.
For example, Mellinger~\textit{et al}.~\cite{2011IROS_DMECAGM_Mellinger} and Lee~\textit{et al}.~\cite{2015CASE_AMUnknownPayload_Lee, 2017TIE_ECPAAT_Lee} estimate unknown payloads by identifying changes in dynamics post-grasping, but  require over 20 s to converge.
Kalman filter-based methods~\cite{2019ICRA_Online_Wuest, 2020RAL_IMUBased_Svacha} fuse IMU and motor data but suffer from slow convergence (tens of seconds) and limited accuracy~\cite{2020POGR_ObservabilityAware_Bohm, 2021RSS_FilterBased_Bohm}, while also neglecting manipulator-payload coupling~\cite{2018RAL_Review_Ruggiero}.
Recent work~\cite{2023CDSR_TimeVaryingInertia_Park} includes manipulator variation in estimation but remains in simulation.
Recently, learning-based methods~\cite{2024JGCD_MetaAdaptiveControl_Gao} emerged to leverage neural networks to infer parameters without modeling.
However, all above methods estimate the inertial parameters of the unknown object post-grasping, leading to estimation delays. 
Moreover, they lack knowledge of the object and its properties.
In contrast, we focus on estimating the individual inertial parameters of the payload, since the inertial properties of the aerial platform are already known, which slightly differs from~\cite{2019ICRA_Online_Wuest,2020RAL_IMUBased_Svacha,2017TIE_ECPAAT_Lee} that estimate combined system or payload parameters.
Another challenge is the design of an effective control strategy for AMs operating with a morphing manipulator and various payloads. 
Adaptive control techniques like sliding mode~\cite{2013IROS_AMQuadrotor2dofArm_Kim} or model reference adaptive controllers~\cite{2020ACC_Adaptive_Baraban} are used to handle parameter uncertainty.
Yet, these systems without accelerometer measurements may respond more slowly to changes. 
To achieve high real-time responsiveness, Gain Scheduled (GS) control has shown promise on AMs~\cite{2014JIRS_HybridAdaptiveControl_Orsag, 2022IJDC_Modeling_Coulombe}, with some combining PID and Lyapunov adaptive control~\cite{2014JIRS_HybridAdaptiveControl_Orsag} to address changes in the CoM and external disturbances. 
However, existing designs that rely only on control error make it sensitive to noise or disturbances~\cite{2017TIE_ECPAAT_Lee}. 
Most current GS-based controllers~\cite{2014JIRS_HybridAdaptiveControl_Orsag, 2022IJDC_Modeling_Coulombe} ignore payload variations, focusing solely on manipulator morphing.

\begin{figure*}[t]
\centering
\includegraphics[width=1.9\columnwidth]{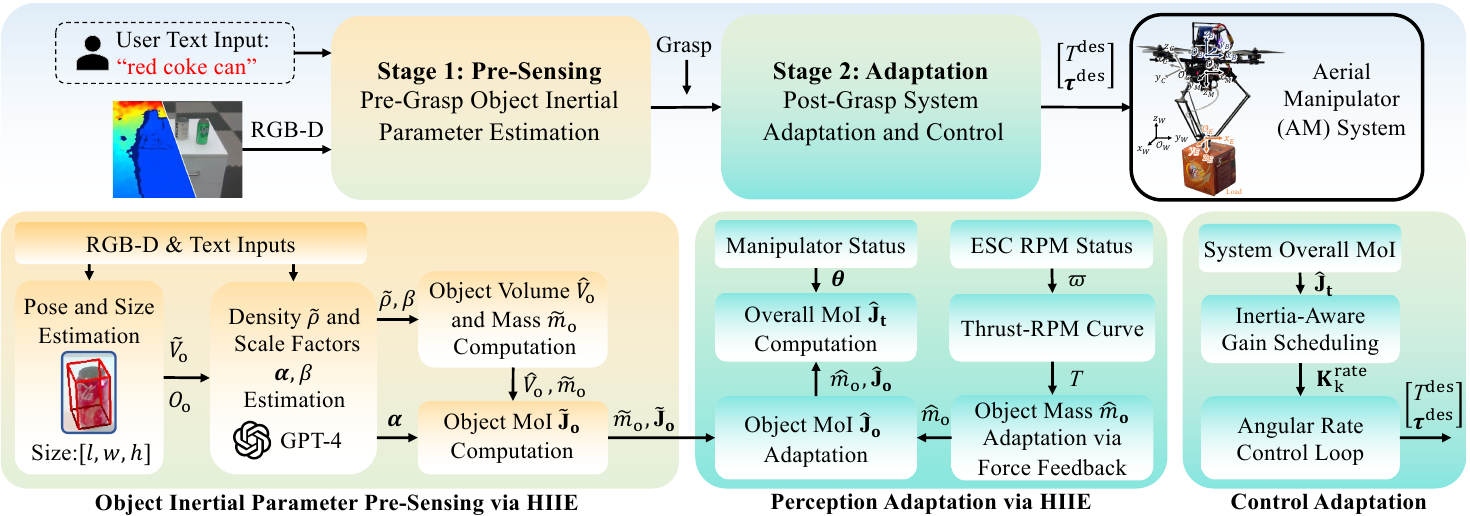}
\caption{
\textbf{System overview.}
Our system processes two inputs: an RGB-D image stream of the scene and text specifying the target object. 
\textbf{Stage 1: Pre-Sensing:} 
Before grasping, the unknown target object's inertial parameters (mass ${\tilde{m}}_o$ and MoI $\tilde{\mathbf{J}}_o$) are pre-estimated by large vision and language models, 
and the object frame origin $\mathcal{O}_O$ is also obtained.
\textbf{Stage 2: Adaptation:} 
Upon grasping, the \textit{Perception Adaptation} module updates the object's mass and MoI to ${\hat{m}}_o$ and $\hat{\mathbf{J}}_o$ using onboard force sensor feedback, then computes the system's total MoI $\hat{\mathbf{J}}_t$. 
The \textit{Control Adaptation} module's inertia-aware gain scheduling (IAGS) controller then calculates the inertia-aware gain, generating desired thrust magnitude $T^{\mathrm{des}}$ and torque $\bm{\tau}^{\mathrm{des}}$.}
\label{fig:system_overview}
\vspace{-0.6cm}
\end{figure*}
Interestingly, humans approach grasping in a two-stage manner. First, they visually estimate object properties—such as size, shape, or texture—and integrate prior experience to infer mass and inertia before physical interaction~\cite{2015FHN_SensorimotorMemory_Polanen}. Second, after contact, tactile feedback allows rapid refinement and force adjustment to ensure stability~\cite{1988EBR_Coordinated_Johansson, 2024SA_FastGripForce_Delhaye}. This preemptive-inference and feedback-refinement loop enables fast and robust grasping—unlike most existing AM systems, which rely on post-grasp estimation and reactive control.
Motivated by the human two-stage grasping strategy, we propose a perception-control framework for aerial manipulation (Fig.~\ref{fig:system_overview}) that mirrors this process:

\textbf{Stage 1: Pre-Sensing}: Before grasping, the system visually perceives the target object using RGB-D images and natural language input. It estimates the object’s pose, size, and shape, then infers physical priors such as density and volume to approximate its mass and moment of inertia. This stage mimics the way humans form expectations about an object’s weight and stability through sight and experience. In robotic systems, this requires visual-semantic understanding and 3D reasoning under uncertainty.
Compared to traditional methods it avoids the need for time-consuming maneuvers like hovering or active excitation~\cite{2019ICRA_Online_Wuest, 2017TIE_ECPAAT_Lee}, and removes dependencies on known object models or manual data annotation~\cite{2018RAL_IntegratedCooperative_Lee, 2015NIPS_Galileo_Wu, 2017CoRL_image2mass_Standley}.

\textbf{Stage 2: Adaptation}: Once the object is grasped, the system refines its estimates of mass and MoI using onboard sensor feedback. These updated inertial parameters are then used to adjust the control system in real time. This stage ensures robust, responsive flight performance under changing load conditions and manipulator configurations.
Specifically, we design an adaptive control strategy that dynamically adjusts control gains based on estimated inertia, without requiring predefined gain maps or operating points. This allows the system to maintain precise and agile control despite payload variations and structural reconfigurations.

Our key contributions are as follows:
\begin{enumerate}
\item \textbf{A Novel Inertia Estimation Framework:}
We design a two-stage framework that combines vision-based pre-grasp estimation with post-grasp adaptation to enable real-time inertial parameter updates for aerial manipulators.

\item \textbf{An Inertia-Compensated Adaptive Strategy:}
We develop a gain-scheduled adaptive controller that compensates for inertia variations, and validate its robustness through frequency-domain system identification.

\item \textbf{Successful Implementation and Verification:}

We demonstrate the versatility and effectiveness of our framework through extensive validation, including diverse tasks, ablation studies, and real-world aerial manipulation experiments.
To the best of our knowledge, this is the first introduction of vision-based inertia estimation techniques to aerial manipulators.
\end{enumerate}

\section{SYSTEM MODELING}
\label{sec:SYSTEM MODELING}
\subsection{AM System Configuration}
\label{subsec:AM_System_Configuration}
Our AM system consists of a quadrotor as a flying base and a delta manipulator. 
For modeling, we set up five coordinate systems: the world frame $\mathcal{F}_W$, the body frame $\mathcal{F}_B$, the manipulator frame $\mathcal{F}_M$, the end-effector frame $\mathcal{F}_E$, and the camera frame $\mathcal{F}_C$, each with a set of orthogonal bases.    
The origins $\mathcal{O}_B$, $\mathcal{O}_C$, $\mathcal{O}_M$, and $\mathcal{O}_E$ are the CoM of the AM, camera optical center, midpoint of three manipulator servos, and geometric center of the end-effector plane, respectively.

\subsection{AM System Dynamics}
\label{subsec:System Dynamics}
The linear and angular dynamics of the AM system  can be summarized by the following equations:
\begin{subequations}
\begin{align}
\bm{\tau}(\bm{\theta}) & = \textstyle\sum_{i=1}^4 \left(\triangle \bm{c}_{i} (\bm{\theta}) \times \bm{T}_i  + \bm{\tau}_i \right),
\label{eq:tau_theta} 
\\
m_t{}^W \mathbf{\dot{v}}_B & = -m_t g \bm{e}_3 + {}^W \mathbf{R}_B \textstyle\sum_{i=1}^4 \bm{T}_i, 
\label{eq:linear_acceleration} 
\\
\mathbf{J}_{t}{}^B(\bm{\theta}) \bm{\dot{\omega}}_B & = \bm{\tau}(\bm{\theta}) - {}^B \bm{\omega}_B \times \mathbf{J}_{t}{}^B(\bm{\theta}) \bm{\omega}_B, \label{eq:angular_acceleration} 
\\
{}^W \mathbf{\dot{R}}_B & = {}^W \mathbf{R}_B {}^B \bm{\hat{\omega}}_B. 
\label{eq:rotation_matrix}
\end{align}
\end{subequations}

where 
$\bm{\theta}=[\theta_1,\theta_2,\theta_3]^\intercal \in\mathbb{R}^3$ is the manipulator's joint angles, controlled by three servos.  
$\triangle \bm{c}_{i}(\bm{\theta})\in\mathbb{R}^3$ represents the distance from each propeller center to the AM CoM. 
$\bm{T}_i \in\mathbb{R}^3$ and $\bm{\tau}_i \in\mathbb{R}^3$ are the force and torque generated by each propeller. 
$m_t$ is the total mass of the AM. 
${}^W\mathbf{v}_B \in\mathbb{R}^3$ is the linear velocity of $\mathcal{F}_B$ with respect to $\mathcal{F}_W$. 
$g$ is the gravitational acceleration, and $\bm{e}_3=[0, 0, 1]^\intercal$ in $\mathcal{F}_W$. 
${}^W\mathbf{R}_B \in\mathbb{R}^{3\times3}$ is the rotation of $\mathcal{F}_B$ relative to $\mathcal{F}_W$. 
${}^B\bm{\omega}_B \in\mathbb{R}^3$ is the angular velocity in $\mathcal{F}_B$. 
$\mathbf{J}_{t}\in\mathbb{R}^{3\times3}$ is the total inertia of the AM in $\mathcal{F}_B$ which is both a function of the angle $\bm{\theta}$ and time. 
$\hat{(\cdot)}$ denotes the skew-symmetric operator.

\subsection{Delta Manipulator Kinematics}
\label{subsec:Delta Manipulator Kinematic}

The 3-DOF parallel delta manipulator with 3-RSS (Revolute-Spherical-Spherical)  configuration similar to~\cite{2025CoRR_WholeBody_Deng} is favored for its compact design and low MoI, ideal for rapid and accurate position control~\cite{2022TRO_PPFARM_Ollero}.
Codourey~\cite{1996IROS_DELTA_Codourey} describes the forward kinematics of the manipulator, where the mapping $h_E: \mathbb{R}^3\to\mathbb{R}^3$ transforms the manipulator's joint angles $\bm{\theta}$ into the translation vector ${}^M\bm{p}_E$ from the manipulator frame origin $\mathcal{O}_M$ to the end-effector frame origin $\mathcal{O}_E$. 
Guglielmetti et al.~\cite{1994IFAC_ClosedForm_Guglielmetti} derive the differential kinematics, computing the end-effector velocity Jacobian $\bm{\mathcal{J}}$ to relate the joint velocities $\dot{\bm{\theta}}$ to the end-effector velocity ${}^M\mathbf{v}_E$.
These relationships can be expressed as:
\begin{subequations}
\begin{align}
{}^M\bm{p}_E &= h_E(\bm{\theta}),
\label{eq:delta_kinematic_forward} 
\\
{}^M\mathbf{v}_E &= \bm{\mathcal{J}} \dot{\bm{\theta}}.
\label{eq:delta_kinematic_differential}
\end{align}
\end{subequations}

Given the desired end-effector's position ${}^M\bm{p}_E$ and velocity ${}^M\mathbf{v}_E$, the required joint states $\bm{\theta}$ and $\dot{\bm{\theta}}$ can be computed via inverse kinematics from Eqs.~\eqref{eq:delta_kinematic_forward} and \eqref{eq:delta_kinematic_differential}. 

\section{INERTIA PARAMETERS ESTIMATOR}
\label{sec:INERTIAL PARAMETER ESTIMATOR}
In contrast to existing methods~\cite{2011IROS_DMECAGM_Mellinger, 2015CASE_AMUnknownPayload_Lee, 2017TIE_ECPAAT_Lee, 2019ICRA_Online_Wuest, 2020RAL_IMUBased_Svacha, 2023CDSR_TimeVaryingInertia_Park, 2024JGCD_MetaAdaptiveControl_Gao} that involve time-consuming post-grasp estimation of the target object's inertial parameters, we propose a more efficient human-inspired strategy. 
We pre-sense and pre-estimate the object's inertial parameters, followed by an onboard adaptation of these estimate.
Subsequently, we compute the overall system's inertial parameters, encompassing both the AM and the target object as the payload.

\subsection{Pre-Grasp Target Object Pre-Sensing}
\label{subsec:Object Pre-Sensing}

The target object for grasping is specified by the user through text input when the AM initiates the mission.
During Pre-Sensing, our method detects the target object in RGB-D images, estimates its pose and size, and simultaneously determines its shape scale factors and density using pre-trained vision and language models~\cite{2024CoRR_GroundedSAM_Ren, 2023ICCV_ISTNet_Liu, 2023CoRR_GPT4_OpenAI}.
The object's inertial parameters are then computed from these model.

\subsubsection{Object Detection}
\label{subsubsec:Object Detection}
To detect unseen target objects without per-instance
training, we resort to Grounded~SAM~\cite{2024CoRR_GroundedSAM_Ren}.
Given the text prompt, the model generates a binary mask.
Because the detector is fully foundation-model based, no
dataset or fine-tuning is required for new objects.

\subsubsection{Object Pose and Volume Estimation} 
\label{subsubsec:Object Pose and Volume Estimation}

\paragraph*{Back-projection and point-set formation} The binary mask is applied to the depth map to lift a target-only point cloud. Because an unoriented bounding box can severely over-estimate volume when the object is rotated, we cast the task as a 9D pose problem and recover an oriented, tight box.

\paragraph*{9D Object Pose Estimation}

\begin{figure}[t]
\centering
\includegraphics[width=.99\columnwidth]{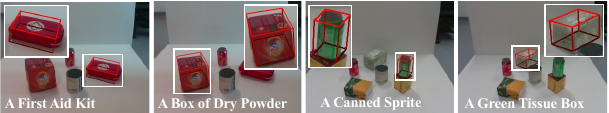}
\caption{Pose estimation results in the camera coordinate frame.}
\vspace{-.4cm}
\label{fig:6object}
\end{figure}

For object pose and size estimation, we adopt IST-Net~\cite{2023ICCV_ISTNet_Liu}, a state-of-the-art category-level method for 6D pose (rotation and translation) and 3D size estimation, resulting in a 9D representation. 
In our AM setting, IST-Net provides reliable estimates at stand-off distances of \SI{3}{–}\SI{5}{m}, and thus serves as the core perception module in our framework.
Multiple tests on 8 objects averaging an average  size estimation accuracy of $90.6\%$. 
Some pose estimation results are shown in Fig.~\ref{fig:6object}.

\subsubsection{GPT-Based Scale Factor and Density Estimation}
\label{subsubsec:GPT-Based Scale Factor and Density Estimation}

\begin{figure}[t]
\centering
\includegraphics[width=0.99\columnwidth]{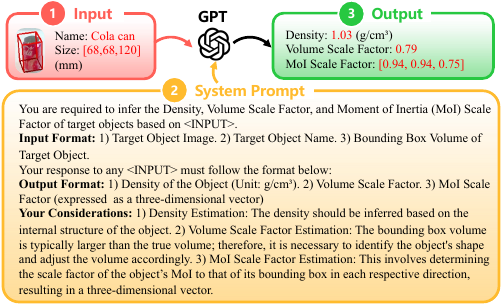}
\caption{System prompt for object scale factor and density estimation.}
\vspace{-0.6cm}
\label{fig:gpt_example}
\end{figure}

Although pose estimation yields tight 3D bounding boxes, real-world objects often deviate from box-like geometries (e.g., spheres, cones), leading to inaccurate volume and inertia estimates.
To address this, we leverage GPT’s multimodal reasoning capabilities to refine volume and MoI estimates using object-specific prior knowledge (Fig.~\ref{fig:gpt_example}). For real-time deployment, the GPT API is accessed onboard via an internet connection, enabling fast and accurate inference.

GPT takes as input a segmented image, approximate object dimensions from point cloud data, and a textual description of the object (e.g., “a red cylindrical iron can filled with coffee beans”).
We use a standardized prompt template (Fig.~\ref{fig:gpt_example}) to ensure reproducibility. GPT predicts:
	1.	a volume scaling factor $\beta$,
	2.	a diagonal MoI scaling matrix $\bm{\alpha} \in \mathbb{R}^{3 \times 3}$, and
	3.	the object’s average density $\tilde{\rho}$.
These are used to compute the refined object volume $\hat{V}_o$, mass $\tilde{m}_o$, and MoI $\tilde{\mathbf{J}}_o$:
\begin{subequations}
\begin{align}
        \label{eq:inertia_and_volume_estimate}
        \tilde{m}_o &= \tilde{\rho} \hat{V}_o = \tilde{\rho} \beta \tilde{V}_o, 
        \\
        \tilde{\mathbf{J}}_o &= \frac{\bm{\alpha}}{12}\tilde{m}_o\cdot
        \text{diag}(w^2 + h^2,\ell^2 + h^2,\ell^2 + w^2).
\end{align}
\end{subequations}

This enables zero-shot generalization to unseen object types, making it suitable for open-set aerial manipulation.

\subsection{Post-Grasp Target Object Inertia Adaptation}
\label{subsec:Mass and Moment of Inertia Update}
To further refine the vision-based estimates of mass $\tilde{m}_o$ and MoI $\tilde{\mathbf{J}}_o$, we employ a disturbance observer (DOB)~\cite{2024IROS_DOBbased_Yu} to adaptively update the mass and scale the MoI accordingly.

\subsubsection{Object Mass Adaptation} 
\label{subsubsec:Object Mass Adaptation}
Upon grasping the target object, it can be considered as subjected to a constant external force that steadily lifts it while maintaining a level attitude. 
The grasping event is detected from the DOB-estimated external force, where the threshold is set above the hover noise level yet below the steady force of the lightest payload, and the condition must persist for at least $0.5$~s before switching.
Once triggered, this allows us to update the object's mass estimation using onboard sensor measurements. 
Inspired by the DOB framework in~\cite{2024IROS_DOBbased_Yu}, we adopt the following approach, which we refer to as DOBm for clarity, distinguishing it from the conventional use of DOB for direct disturbance compensation:

\begin{equation}
\dot{\hat{m}}_o g \bm{e}_3 = \frac{c}{m_a} ( m_a {}^W\mathbf{a}_B - m_a g \bm{e}_3 + {}^W\mathbf{R}_B \bm{T} - \hat{m}_o g \bm{e}_3 ).
\label{eq:mass_update_continuous} 
\end{equation}

Here
$\hat{m}_o$ is the estimated mass of the target object; 
$m_a$ is the AM's mass in its unloaded state; 
$c$ is a constant convergence parameter.
${}^W \mathbf{a}_B$ is the linear acceleration of the flying base in $\mathcal{F}_W$, derived from the IMU of the flight controller in real time;
$\bm{T}$ is the total thrust, computed as $\bm{T}=\sum_{i=1}^4 \bm{T}_i$, where $\bm{T}_i$ is the thrust from the $i$-th propeller.
For each propeller, the thrust magnitude $T_i = ||\bm{T}_i||$ is proportional to the square of its motor's rotation rate $\varpi_i$ (in RPM) as $T_i = c_T \varpi_i^2$~\cite{2014AIAA_ReynoldsNumberEffects_Deters}, and the thrust coefficient $c_T$ is empirically determined through in-situ calibration under actual flight conditions.
Unlike conventional static bench tests—which often fail to capture the complex aerodynamic interactions present during real flights—our calibration is conducted directly onboard during flight, ensuring higher relevance to operational dynamics.

\subsubsection{Object MoI Adaptation} 
\label{subsubsec:Object MoI Adaptation}
In our framework, the object mass is first estimated from Eq.~(3) as $\tilde{m}_o$, and then refined online by the DOB in Eq.~(4) to obtain $\hat{m}_o$. This correction compensates for inertia deviations caused by inaccurate mass estimation. Using the updated mass $\hat{m}_o$, the object's estimated MoI $\hat{\mathbf{J}}_o$ after adaptation is scaled proportionally from the initial estimate $\tilde{\mathbf{J}}_o$ as:
\begin{equation}
\hat{\mathbf{J}}_o = \tilde{\mathbf{J}}_o(\hat{m}_o/\tilde{m}_o).
\label{eq:MoI_update}
\end{equation}

\subsection{Post-Grasp System Overall Inertia Computation}
\label{subsec:Overall Moment of Inertia Computation}

After grasping, the system comprises three components: the quadrotor flying base, the  manipulator, and the target object payload. 
Since the delta manipulator's lightweight and compact design results in low inertia~\cite{2022TRO_PPFARM_Ollero}, and its movable components account for less than $3\%$ of the overall mass in our system, we neglect the influence of manipulator joint variations on the system's CoM and MoI calculations.

\subsubsection{System Overall Mass} 
\label{subsubsec:System Overall Mass}
The mass of the AM, including the flying base and the manipulator in an unloaded state, is measured as $m_a$, and the target object's mass estimation is updated as $\hat{m}_o$. 
Therefore the system's total estimated mass $\hat{m}_t = m_a + \hat{m}_o$. 

\subsubsection{System Overall CoM} 
\label{subsubsec:System Overall CoM}
The system's CoM is computed as the mass-weighted average of all components (the AM body and object) relative to $\mathcal{O}_M$. 
The real-time CoM $\bm{c}_t$ varies with manipulator joint angles $\bm{\theta}$ as:
\begin{equation}
\label{eq:CoM}
\bm{c}_t(\bm{\theta})=(m_a{}^M\bm{p}_B + \hat{m}_o{}^M\bm{p}_O(\bm{\theta}))/ \hat{m}_t.  
\end{equation}

Here ${}^M\bm{p}_B$ is the vector from $\mathcal{O}_M$ to the unloaded AM's CoM $\mathcal{O}_B$ (measured via bifilar pendulum method~\cite{2009JAircraft_BifilarPendulum_Jardin}).

${}^M\bm{p}_O(\bm{\theta})={}^M\bm{p}_E(\bm{\theta})+{}^E\bm{p}_O$ 
is the vector from $\mathcal{O}_M$ to the object frame origin $\mathcal{O}_O$, assuming $\mathcal{O}_O$ coincides with the object's CoM. 
${}^M\bm{p}_E(\bm{\theta})$ and ${}^E\bm{p}_{O}$ are the vectors from $\mathcal{O}_M$ to $\mathcal{O}_E$, and from $\mathcal{O}_E$ to $\mathcal{O}_O$, respectively.

\subsubsection{System Overall MoI} 
\label{subsubsec:System Overall MoI}
The system's total estimated MoI $\hat{\mathbf{J}}_{t}$ is derived via the parallel axis theorem: 

\begin{equation}
\begin{aligned}
\hat{\mathbf{J}}_{t}
= \mathbf{J}_{a} + \hat{\mathbf{J}}_{o} 
&+ m_{a}\big[(\bm{d}_{a}\cdot\bm{d}_{a})\mathbf{I}_3 - \bm{d}_{a}\otimes\bm{d}_{a}\big] \\
&+ \hat{m}_{o}\big[(\bm{d}_{o}\cdot\bm{d}_{o})\mathbf{I}_3 - \bm{d}_{o}\otimes\bm{d}_{o}\big].
\end{aligned}
\label{eq:J_t}
\end{equation}

Here 
$\mathbf{J}_a$ is the AM’s inertia in its unloaded state (measured via bifilar pendulum~\cite{2009JAircraft_BifilarPendulum_Jardin}). 
$\hat{\mathbf{J}}_o$ is the object's inertia from Eq.~\eqref{eq:MoI_update}. 
$\bm{d}_a$ and $\bm{d}_o$ are the vectors from the unloaded AM's CoM $\mathcal{O}_B$, and the object's CoM $\mathcal{O}_O$, to the system's total CoM $\bm{c}_t(\bm{\theta})$, respectively. 
$\mathbf{I}_3$ is the $3\times3$ identity matrix. 
$\otimes$ denotes the outer product.

\section{CONTROLLER DESIGN}
\label{sec:CONTROLLER DESIGN}
We adopt a decoupled control approach~\cite{2018RAL_Review_Ruggiero, 2022TRO_PPFARM_Ollero} to separately control the quadrotor flying base and the delta manipulator as two subsystems.
This design choice is motivated by the delta manipulator’s lightweight and low-inertia structure, which introduces minimal dynamic coupling during typical aerial operations.
While we neglect the manipulator’s mass and inertia in the control formulation, the payload is explicitly considered during both the inertia estimation and adaptive gain scheduling stages (Eq.~(\ref{eq:J_t})).
As a result, dominant coupling effects arising from the grasped object are accounted for, while residual coupling from the arm remains negligible under our experimental conditions, consistent with prior work~\cite{2022TRO_PPFARM_Ollero}.

\begin{figure}[t]
\centering
\includegraphics[width=0.99\columnwidth]{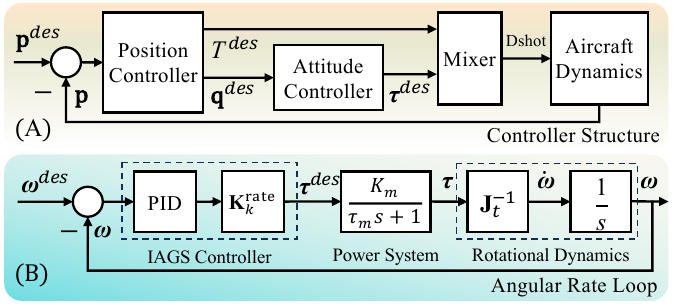}
\caption{Inertia-compensated adaptive strategy.
\textbf{(A) Quadrotor Control Structure}:
Comprises a position controller, an attitude controller, and a mixer. 
\textbf{(B) Angular Rate Control Loop}:
The PID control gains $\mathbf{K}_k^{\text{rate}}$ are dynamically adjusted through Inertia-Aware Gain Scheduling (IAGS) based on the manipulator joint angles $\bm{\theta}$.}
\label{fig:adaptive_control}
\vspace{-0.5cm}
\end{figure}

\subsection{Control of the Delta Manipulator}
\label{subsec:Delta Manipulator Control}

The delta manipulator's controller tracks predefined end-effector trajectories (desired position ${}^M\bm{p}_E^{\text{des}}$ and velocity ${}^M\mathbf{v}_E^{\text{des}}$) by commanding desired joint angles $\bm{\theta}^{\text{des}}$ and angular velocities $\bm{\dot{\theta}}^{\text{des}}$ to the servos:
\begin{subequations}
\begin{align}
\bm{\theta}^{\text{des}} &= h_E^{-1}({}^M\bm{p}_E^{\text{des}}),  
\\
\bm{\dot{\theta}}^{\text{des}} &= \mathbf{\mathcal{J}^{-1}} {}^M\mathbf{v}_E^{\text{des}}+\mathbf{K}_{\theta}(\bm{\theta}^{\text{des}}-\bm{\theta}).  
\label{eq:delta_kinematic}
\end{align}
\end{subequations}

Here
$\bm{\theta}^{\text{des}}$ is computed via inverse kinematics of Eq.~\eqref{eq:delta_kinematic_forward}. 
$\bm{\dot{\theta}}^{\text{des}}$ 
combines feedforward velocity derived from inverse kinematics of Eq.~\eqref{eq:delta_kinematic_differential} with proportional (P) control feedback. 
$\mathbf{K}_{\theta}$ is a positive definite gain matrix for the proportional control. 
$\bm\theta$ is the real-time joint angles sampled at 100 Hz.

\subsection{Control of the Quadrotor Flying Base}
\label{subsec:Flying Base Control}

\subsubsection{Differential-Flatness-Based Cascade Feedback Control}
Fig.~\ref{fig:adaptive_control} (A) illustrates our overall control strategy, which adopts a widely used cascade feedback architecture consisting of a position controller, an attitude controller, and a mixer. The position controller takes the desired position $\mathbf{p}^{des}$ as input and generates the desired thrust $T^{des}$ and desired attitude $\mathbf{q}^{des}$ based on the differential flatness properties of quadrotors~\cite{2025CoRR_NDOBbased_Chen}. The attitude controller then outputs the desired torque $\boldsymbol{\tau}^{des}$. 
The attitude error is defined in a geometric manner on $SO(3)$, following standard formulations in~\cite{lyu2017design}.

\subsubsection{Angular Rate Loop with Inertia-Aware Gain Scheduling (IAGS)}
- \textit{IAGS Controller Design}. 
The controller in the angular rate loop (Fig.~\ref{fig:adaptive_control}~(B)) takes the angular rate error $\bm{\omega}^{\text{des}}-\bm{\omega}$ as input and outputs the desired torque $\bm{\tau}^{\text{des}}$. 
It employs a standard PID controller as its backbone, with the proportional, integral, and derivative gains $\mathbf{K}_p^{\text{rate}}, \mathbf{K}_i^{\text{rate}}, \mathbf{K}_d^{\text{rate}} \in \mathbb{R}^{3\times3}$ determined through parameter tuning.
To counteract MoI variations induced by the manipulator's motion, we introduce an additional inertia-aware adaptive term $\mathbf{K}_k^{\text{rate}}$.
For clarity, the derivative term is expressed as the ideal operator $s$ in the frequency-domain analysis, while in practice it is implemented via a discrete-time causal approximation.
Thus, the controller transfer function in the Laplace domain is expressed as:
\begin{align}
    G_{c}(s) = \left(\mathbf{K}_p^{\text{rate}}+\mathbf{K}_i^{\text{rate}}\frac{1}{s}+\mathbf{K}_d^{\text{rate}}s\right)\mathbf{K}_k^{\text{rate}}. 
\label{eq:G_c}
\end{align}
To facilitate implementation, the corresponding time-domain control law is given by:
\begin{align}
    \bm{\tau}_T^{des} =\; \Big( \mathbf{K}_p^{\text{rate}}\,\bm{\omega}^e 
    \;+\; \mathbf{K}_i^{\text{rate}} \!\int \bm{\omega}^e \, dt
    \;+\; \mathbf{K}_d^{\text{rate}} \frac{d \bm{\omega}^e}{dt} \Big)\, \mathbf{K}_k^{\text{rate}},
\label{eq:G_c_t}
\end{align}
where $\bm{\tau}_T^{des}$ denotes the control torque output in the time domain, given the angular rate error $\bm{\omega}^e$.

\noindent
- \textit{Power System}. 
The controller output $\bm{\tau}^{\text{des}}$ is fed into the Mixer to generate the rotational speed. 
The rotational speed signal is then fed to the ESC, which controls the motor and propeller to produce the actual torque $\bm{\tau}$. 
The transfer function from $\bm{\tau}^{\text{des}}$ to $\bm{\tau}$ can be written as:
\begin{equation} 
G_m(s) = \frac{K_m}{\tau_m s + 1}, 
\label{eq:G_m}
\end{equation}
where 
$K_m\in\mathbb{R}$ is the steady-state gain and 
$\tau_m$ is the average time constant.

\noindent
- \textit{Rotational Dynamics}. 
The control torque $\boldsymbol{\tau}$ acts on the rotational dynamics, generating the corresponding angular velocity $\boldsymbol{\omega}$. 
This approximation is justified by a small-perturbation linearization about hover $(\boldsymbol{\omega}^\star = \mathbf{0})$, where the quadratic gyroscopic term is negligible and $\mathbf{J}_t$ can be treated as locally diagonal. This matches our operating condition, as all grasping experiments are conducted from hover.
Under these assumptions, the transfer function is:
\begin{equation} 
G_d(s) = \mathbf{J}_t^{-1}\frac{1}{s}, 
\label{eq:G_d}
\end{equation}
where
$\mathbf{J}_t$ is the actual total MoI of the AM system.

\noindent
- \textit{Angular Rate Loop Transfer Function}. 
The overall open-loop transfer function of the angular rate loop, $G_{\text{open}}^{\text{rate}}(s)$, is: 
\begin{equation}
\begin{aligned}
G_{\text{open}}^{\text{rate}}(s) 
&= G_c(s) G_m(s) G_d(s) \\
&= \frac{K_m \left(\mathbf{K}_p^{\text{rate}}s + \mathbf{K}_i^{\text{rate}} + \mathbf{K}_d^{\text{rate}}s^2\right) 
       \mathbf{K}_k^{\text{rate}} \mathbf{J}_t^{-1}}{s^2(\tau_m s + 1)}.
\label{eq:G_rate} 
\end{aligned}
\end{equation}

\noindent
- \textit{Determine Inertia-Aware Adaptive Gain $\mathbf{K}_k^{\text{rate}}$}. 
We therefore design the inertia-aware adaptive gain $\mathbf{K}_k^{\text{rate}}$ as:
\begin{equation}
\mathbf{K}_k^{\text{rate}} = \mathbf{J}_a^{-1} \hat{\mathbf{J}}_{t}.  
\label{eq:K_k}
\end{equation}

\begin{figure}[t]
\centering
\includegraphics[width=0.9\columnwidth]{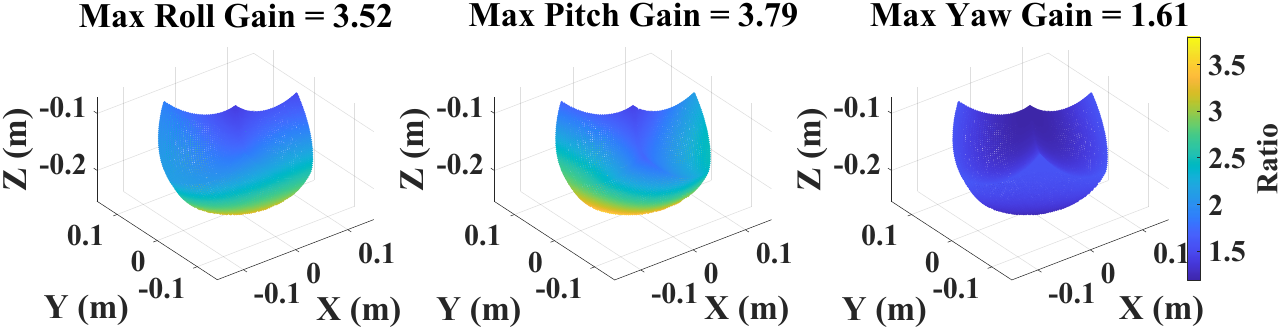}
\caption{The maximum values of the gain $\mathbf{K}_k^{\text{rate}}$ along the three directions (x, y, and z).}
\label{fig:Workspace_K}
\vspace{-0.6cm}
\end{figure}

The determination of $\mathbf{K}_k^{\text{rate}}\in \mathbb{R}^{3\times3}$ is crucial for achieving stable control.
This gain is designed to dynamically adjust with changes in the system's MoI $\mathbf{J}_{t}$, ensuring consistent control performance across various load conditions.
The reasons and advantages for this design are as follows:

\noindent
- The controller gain $\mathbf{K}_k^{\text{rate}}$ dynamically adjusts with changes in the system's MoI $\mathbf{J}_t$. 
When $\mathbf{J}_t$ increases (e.g., due to manipulator motion or added payload), $\mathbf{K}_k^{\text{rate}}$ increases to compensate for the higher inertia. 
When $\mathbf{J}_t$ decreases, $\mathbf{K}_k^{\text{rate}}$ decreases to avoid over-response. 

\noindent
- As detailed in Section~\ref{subsec:Overall Moment of Inertia Computation}, the total MoI $\mathbf{J}_t$ of the AM depends on $\mathbf{J}_a$, $\mathbf{J}_o$, and $\bm{\theta}$. 
When the system is in its unloaded state, $\mathbf{J}_t = \mathbf{J}_a$, and thus $\mathbf{K}_k^{\text{rate}} \approx \mathbf{I}_3$ (the identity matrix). 
This ensures that the additional term $\mathbf{K}_k^{\text{rate}}$ does not affect the control of the AM in the unloaded state. 

\noindent
- Substituting Eq.~\eqref{eq:K_k} into Eq.~\eqref{eq:G_rate}, and noting that $\hat{\mathbf{J}}_t \mathbf{J}_t^{-1} \approx \mathbf{I}_3$, the open-loop transfer function $G_{\text{open}}^{\text{rate}}(s)$ becomes independent of $\mathbf{J}_t$. 
As a result, the system's dynamic characteristics are primarily determined by the nominal MoI $\mathbf{J}_a$, rather than the current MoI $\mathbf{J}_t$. 
This ensures consistent control performance across various load conditions, allowing the system to remain stable even during manipulator motion or changes in payload. 

\noindent
- This design is simple yet effective, reducing the complexity of parameter tuning. 
Computationally, $\mathbf{J}_a$ is preconfigured, and $\hat{\mathbf{J}}_t$ can be quickly updated after grasping the  object using Eq.~\eqref{eq:J_t}, enabling real-time adaptation to changes in MoI. 

\begin{figure}[t]
\centering
\includegraphics[width=0.85\columnwidth]{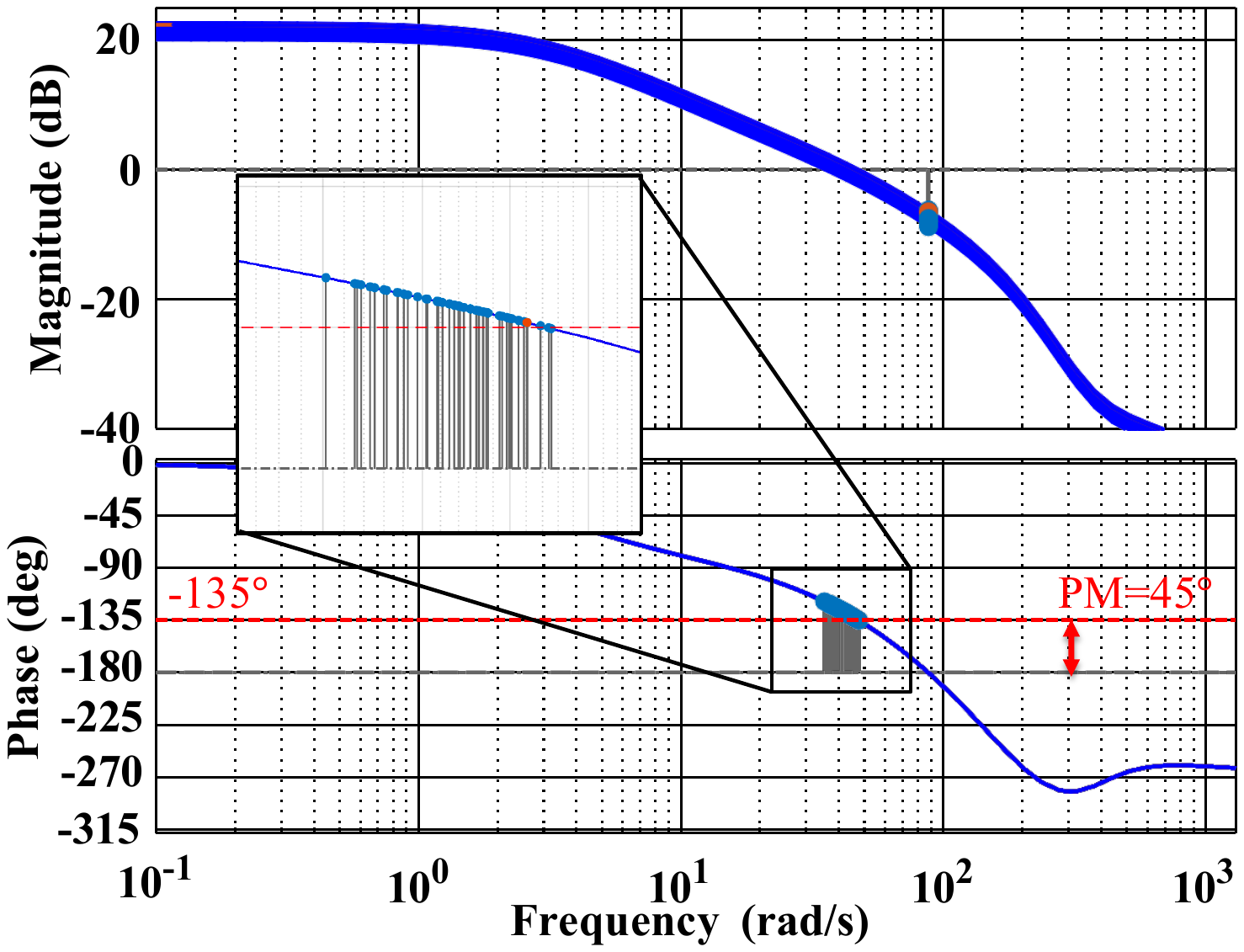}
\caption{Open-loop bode plot with uncertainty.}
\label{fig:OL_margin}
\vspace{-0.5cm}
\end{figure}
\subsubsection{Frequency-Domain Stability Assessment}
\label{sec:Frequency-Domain Stability Assessment}

To evaluate closed-loop stability under payload-induced disturbances, we perform a $\mu$-analysis that jointly accounts for uncertainties in the post-grasp inertia and in the scheduled rate gain. The uncertainty bounds are physically grounded in the system’s payload envelope: we assume the heaviest admissible object, limited to 400 g within a $20{\times}20{\times}20~\text{cm}^3$ bounding volume. By sweeping the Delta arm across its workspace and recomputing the whole-system inertia about the updated center of mass, this payload produces worst-case inertia scaling factors of $3.52\times$, $3.79\times$, and $1.61\times$ along the $x$-, $y$-, and $z$-axes, respectively (Fig.~\ref{fig:Workspace_K}). In the robustness study, we therefore co-vary the inertia and the scheduled rate gain over the \emph{same per-axis bounds}—anchored at unity and capped at $3.52\times$ (x), $3.79\times$ (y), and $1.61\times$ (z). 
The resulting $\mu$ remains strictly below unity across $[1,600]$~rad/s, confirming robust stability; the $\mu$-critical case still exhibits a $52.4^\circ$ phase margin at $41.5$~rad/s (Fig.~\ref{fig:OL_margin}), providing headroom beyond $45^\circ$.

Finally, we corroborated the analysis in both simulation and flight. Under the worst-case inertia, we swept the scheduled rate gains up to their per-axis bounds—$3.52\times$ (x), $3.79\times$ (y), and $1.61\times$ (z)—and observed no oscillation or divergence. These results support the fidelity of the identified plant and the validity of our stability assessment.

\vspace{-.1cm}

\begin{table}[h]
\centering
\caption{Experimental Setup Parameters}
\renewcommand{\arraystretch}{1.12}
\resizebox{\linewidth}{!}{%
\begin{tabular}{l l | l l}
\toprule
\multicolumn{4}{c}{\textbf{Real-World Experimental Parameters}} \\
\midrule
Mass $m$  & $1.379$ kg &
Pos. gain $\mathbf{K}_{\theta}$ & $(20,20,20)^\dagger$ \\

Inertia $\mathbf{J}_a ({kg\cdot m^2})$ &
$(9.2,10.5,14.7)^\dagger\cdot10^{-3}$ &
Pos. gain $\mathbf{K}^{\text{pos}}$ & $(4,4,3)^\dagger$ \\

Thrust coeff. $c_T$ & $18.1712$ &
Vel. gain $\mathbf{K}^{\text{vel}}$ & $(3.5,3.4,3)^\dagger$ \\

Inner-loop & PX4 v1.14 default, 400 Hz &
DOB gain $c$ & $10$ \\

DOB force LPF & 50\,Hz cutoff & DOB rate & 100\,Hz \\

$\mathcal{L}_1$ moment LPF & 5, 10 rad/s cutoff & $\mathcal{L}_1$ rate & 400\,Hz \\

$\mathcal{L}_1$ Hurwitz gain $\mathbf{A_s}$& $(-10, -10, -10)^\dagger$ & grav. accel. $g$ & $9.81\,\mathrm{m/s^2}$  \\
\bottomrule
\end{tabular}
}
\vspace{2pt}
{\footnotesize %
$^\dagger$Matrices in the table are diagonal. LPF denotes a low-pass filter.}
\label{tab:exp_setup}
\vspace{-.3cm}
\end{table}

\section{EXPERIMENTS AND RESULTS}
\label{sec:EXPERIMENT VERIFICATION}
We evaluate our methods on a custom AM platform weighing 1.379 kg and measuring 240$\times$240$\times$220 mm$^3$.  The platform is equipped with a NxtPX4v2 flight controller, an Intel RealSense D435i depth camera, and an NVIDIA Jetson Orin NX 16 GB for onboard computation.
The delta arm is actuated by three DYNAMIXEL XL430-W250-T servo motors and equipped with an end-effector consisting of three suction cups driven by a vacuum pump.
Fig.~\ref{fig:system_overview} provides an overview of the platform.
Our system is developed in C++11 on Ubuntu 20.04 ROS Noetic with parameters listed in Tab.~\ref{tab:exp_setup}, where $\mathcal{L}_1$Quad~\cite{wu2023mathcal} is applied only to rotational dynamics.
Accurate positioning is provided by the NOKOV motion capture system, while perception, planning, and control modules are executed on the onboard computer.
In our experiments eight different test objects with varying inertial parameters are used: \textit{
1. A solid plastic box, 
2. A yellow tissue box, 
3. A mouse box, 
4. A solid cardboard box, 
5. A canned coffee bean container, 
6. A white iron can with candy, 
7. A dry powder extinguisher box, and 
8. A coke can. }

\begin{table}[t]
\centering
\caption{Inertia Estimation Errors at Two Stages}
\renewcommand{\arraystretch}{1.12}
\resizebox{\linewidth}{!}{%
\small
\renewcommand{\arraystretch}{1.1}
\begin{tabular}{l c c c c c c c}
\toprule
& \bm{$m_t$} & \bm{$c^x_t$} & \bm{$c^y_t$} & \bm{$c^z_t$} & \bm{${J}^{xx}_t$} & \bm{${J}^{yy}_t$} & \bm{${J}^{zz}_t$} \\
& {(kg)} & {(m)} & {(m)} & {(m)} & {$(g\cdot m^2)$} & {$(g\cdot m^2)$} & {$(g\cdot m^2)$} \\
\midrule
$\bm{e}_{\textbf{Pre-Sensing}}$ & 0.0380  & 0.0009  & 0.0002  & 0.0051  & 1.745 & 1.789 & 0.055 \\
$\bm{\delta}_{\textbf{Pre-Sensing}}$ & -2.38\% & -15.33\% & -15.33\% & -21.93\% & -8.49\% & -8.08\% & -0.36\% \\
\midrule
$\bm{e}_{\textbf{Final}}$ & 0.0058  & \text{0.03e-12}  & \text{0.05e-12}  & 0.0003  & 0.051  & 0.051  & 0.001 \\
$\bm{\delta}_{\textbf{Final}}$ & -0.36\% & -2.27\%  & -2.27\%  & 1.98\% & -0.45\% & -0.39\% & \text{-5.19e-3\%} \\
\bottomrule
\end{tabular}}
\vspace{2pt}
{\footnotesize %
$e$ and $\delta$ denote the absolute and relative errors respectively.}
\label{tab:Exp_A_Result_InertiaEstimation}
\vspace{-0.6cm}
\end{table}

\subsection{Inertial Parameter Estimation}
\label{subsec:Inertial Parameter Estimation}

To demonstrate the accuracy and efficiency of our method, we evaluated it on a diverse set of everyday objects of varying shapes, sizes, and masses.
For clarity, we detail here a representative run with a 219 g coffee-can target while the AM flies in, grasps the object, and estimates the following inertial parameters online: the total mass $m_t$, center of mass $\bm{c}_t$, and principal moments of inertia $\mathbf{J}_t$. 
Similar to~\cite{2019ICRA_Online_Wuest}, an estimate is declared converged once it enters and subsequently remains within the following bounds: mass error $<$ 4\%, MoI error $<$ 20\%, and CoM error $<$ 2mm.

\begin{figure}[t]
\centering
\includegraphics[width=0.97\columnwidth]{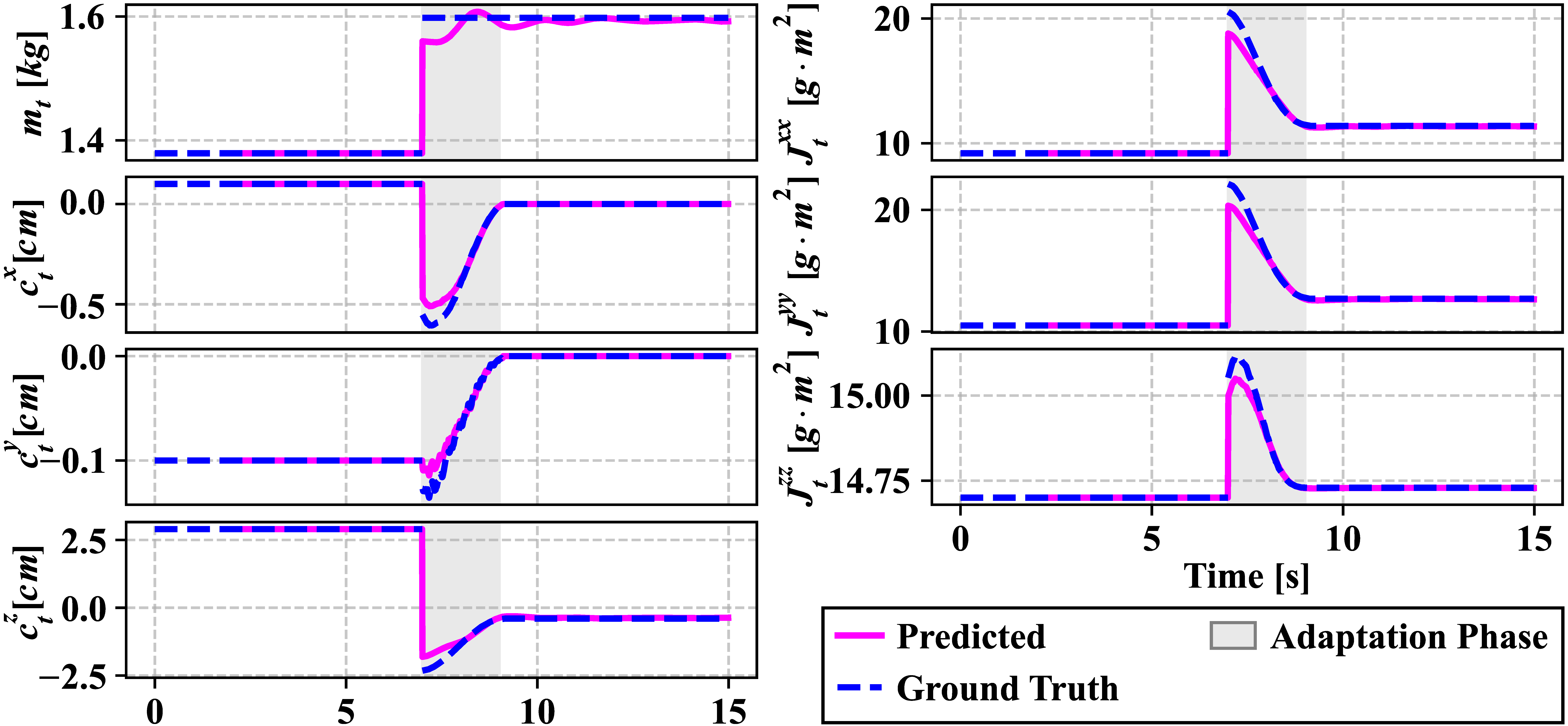}

\caption{
Real-time inertia estimation during object grasping.}
\label{fig:Exp_A_Result_InertiaEstimation}
\vspace{-0.6cm}
\end{figure}

Fig.~\ref{fig:Exp_A_Result_InertiaEstimation} shows real-time estimation results for each parameter. 
At $0~\si{s}$, the AM begins flying toward the target object while performing object Pre-Sensing. 
By $2~\si{s}$, an initial estimate of the object's mass and MoI is obtained, though these values are not yet applied to the system's overall parameters as the object remains ungrasped.
At $7~\si{s}$, the AM successfully grasps the object, triggering an update of the system's inertial parameters to account for the object's influence.
The absolute and relative errors of the estimation results at this Pre-Sensing stage are shown in the first two rows of Tab.~\ref{tab:Exp_A_Result_InertiaEstimation}.  
The relative errors for the total mass, CoM, and MoI are $-2.38\%$, $-21.93\%$ to $-15.33\%$, and $-8.49\%$ to $-0.36\%$, respectively.
Through adaptation based on onboard sensor feedback, the inertial parameters are further refined.
The absolute and relative errors are listed in the last two rows of Tab.~\ref{tab:Exp_A_Result_InertiaEstimation}.
After adaptation, the relative estimation errors for the total mass, CoM, and MoI improve to $0.36\%$, $\leq2.27\%$, and $\leq0.45\%$, respectively.

Compared to~\cite{2019ICRA_Online_Wuest} (errors $<4\%$, $5\%$, $20\%$ for mass, CoM, MoI estimation, respectively) and~\cite{2020RAL_IMUBased_Svacha} (mass error up to $2.2\%$, no CoM/MoI estimation results reported), our Pre-Sensing results are competitive, while the post-adaptation results achieve significantly higher accuracy.
Furthermore, unlike~\cite{2017TIE_ECPAAT_Lee} and~\cite{2019ICRA_Online_Wuest}, which require $27~\si{s}$ of hovering or $19~\si{s}$ of excitation trajectory after grasping, our approach completes the inertia estimation in approximately $2~\si{s}$, achieving an approximately 10-fold improvement in efficiency.

\subsection{Hover Stabilization}
\label{subsec:Hover Stabilization}

\begin{figure}[t]
\centering

\includegraphics[width=0.97\columnwidth]{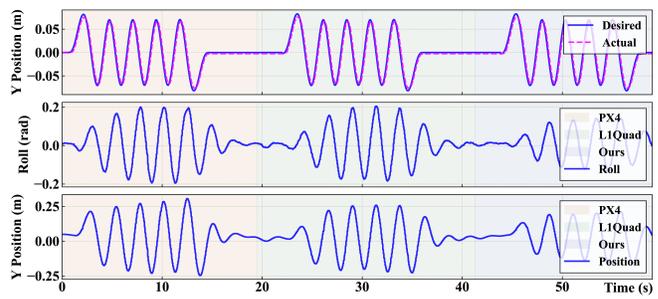}
\caption{End-effector trajectory tracking along the y-axis.}
\label{fig:line}
\vspace{-0.5cm}
\end{figure}

To evaluate the proposed method, we conduct a hover–stabilization benchmark. 
The experiment was conducted with a 400 g iron disc and eight different object mentioned above as the payload. 
Three trajectory segments were executed, each consisting of five back-and-forth motions at 10 cm/s along the body y-axis, under the PX4 v1.14.0~\cite{PX4v114} (0–20 s), the $\mathcal{L}_1$Quad~\cite{wu2023mathcal} (20-40 s), and the proposed method (after 40 s), as shown in Fig.~\ref{fig:line}. 
The top shows the manipulator end-effector $y$-axis tracking, the middle shows the AM roll–attitude response, and the bottom shows the AM $y$-position response. 
Specifically, the PX4 baseline uses the multicopter cascaded attitude–rate PID controller (modules \texttt{mc\_att\_control} and \texttt{mc\_rate\_control}) from PX4 v1.14.0~\cite{PX4v114}. 
We quantify performance using RMSE and maximum absolute error for position; averaged over all eight objects, our method consistently reduces all metrics relative to the baseline. 
As summarized in Tab.~\ref{tab:ypos_table}, both $\mathcal{L}_1$Quad and our method reduce $y$-axis position error under hover disturbance compared to PX4; our controller achieves the largest gains, lowering RMSE by $28.2\%$ and peak error by $30.9\%$ relative to PX4. 
For attitude tracking, our controller likewise improves performance, cutting attitude RMSE by $29.4\%$ and peak attitude error by $28.5\%$ compared with PX4.

\begin{table}[h]
  \caption{Hover–stabilization benchmark}
  \label{tab:ypos_table}
  \centering
  \renewcommand{\arraystretch}{1.12}
  \resizebox{\linewidth}{!}{%
  {
  \begin{tabular}{lcccc}
    \toprule
    \textbf{Method} & \multicolumn{2}{c}{\textbf{Position}} & \multicolumn{2}{c}{\textbf{Attitude (roll)}} \\
    \cmidrule(lr){2-3}\cmidrule(lr){4-5}
           & \textbf{RMSE [m]} & \textbf{Max [m]} & \textbf{RMSE [rad]} & \textbf{Max [rad]} \\
    \midrule
    \textbf{PX4 v1.14}~\cite{PX4v114}  & 0.143$^{\ast}$ & 0.307$^{\ast}$ & 0.099$^{\ast}$ & 0.200$^{\ast}$ \\
    \textbf{$\mathcal{L}_1$Quad}~\cite{wu2023mathcal}     & 0.120 (\(\downarrow\)16.3\%) & 0.259 (\(\downarrow\)15.4\%) & 0.080 (\(\downarrow\)19.3\%) & 0.184 (\(\downarrow\)8.0\%) \\
    \textbf{Ours}       & 0.103 (\(\downarrow\)28.2\%) & 0.212 (\(\downarrow\)30.9\%) & 0.070 (\(\downarrow\)29.4\%) & 0.143 (\(\downarrow\)28.5\%) \\
    \bottomrule
  \end{tabular}
  }
  }
  \vspace{2pt}
  {\footnotesize %
  $^{\ast}$ Reference values. \(\downarrow\) denotes reduction relative to PX4.}
  \vspace{-0.8cm}
\end{table}

\subsection{Object Manipulation and Transportation}
\label{subsec:Manipulation_and_Transportation}

We evaluate our framework in real-world manipulation and transportation scenarios through two representative tasks.
\textbf{Task 1 — Object Pick-and-Place}:
For each of eight test objects, we perform 20 trials using both PX4 and our method. In each trial, the object is randomly placed 3 m away, after which the system detects, grasps, and transports it along a whole-body trajectory~\cite{2025CoRR_WholeBody_Deng} to a table 5 m away at up to 4 m/s (Fig.~\ref{fig:Exp_C2_Scenario_PassGate} (A)). Our method consistently yields lower tracking errors across all objects, with larger gains for heavier payloads. The maximum RMSE improvement reaches 0.054 m, and averaged over all objects, the position-tracking RMSE mean and standard deviation are reduced by $17.47\%$ and $59.91\%$, respectively (Fig.~\ref{fig:Exp_C2_Scenario_PassGate} (D)).
\textbf{Task 2 — Object Transportation Under Disturbances}:
The aerial manipulator transports the object through a 40 cm racing gate while subjected to a lateral 5 m/s wind disturbance (Fig.~\ref{fig:Exp_C2_Scenario_PassGate} (A)). Gate passage requires arm retraction, introducing abrupt inertia changes, and the wind acts 2 m before the gate. Under these challenging conditions, our method reduces position and velocity RMSE by $9.45\%$ and $9.7\%$, and reduces attitude and angular-rate RMSE by $17.25\%$ and $20.35\%$, demonstrating improved robustness during disturbed transportation.

\subsection{Ablation Study}\label{subsec:Ablation_Study}
\begin{figure*}[t]
\centering
\includegraphics[width=1.99\columnwidth] {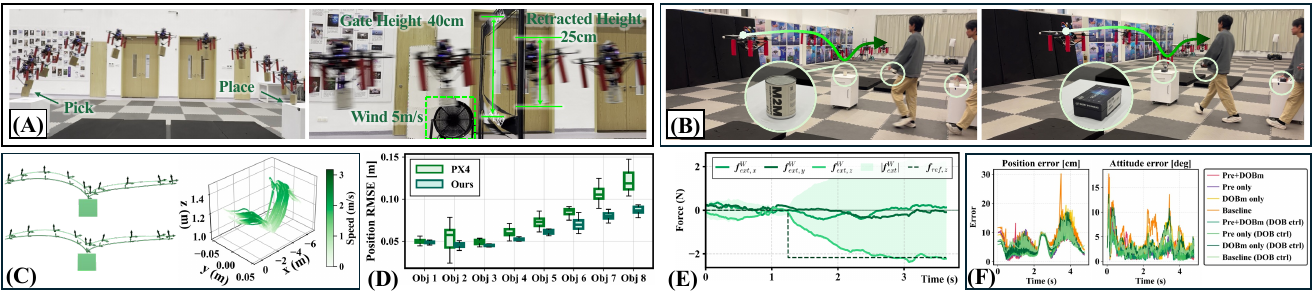}
\caption{(A) Task 1: Object Pick-and-Place (left); Task 2: Object Transportation under Disturbance (right). (B) The AM grasps different object that is randomly placed on the table in the ablation study. (C) Different reference trajectory (left) and 16 flight trials (right) in the ablation study. (D) Position tracking error in Object Pick-and-Place task. (E) DOB estimation of linear external forces during grasping process. (F) Position and attitude errors in the ablation study.}
\label{fig:Exp_C2_Scenario_PassGate} 
\vspace{-.3cm} 
\end{figure*}

This section evaluates the effect of the Pre-Sensing module via ablation experiments in aerial manipulation tasks on two objects (Fig.~\ref{fig:Exp_C2_Scenario_PassGate} (B)).
Two experimental groups are considered: Group1 uses the standard PX4 controller without DOB compensation as the baseline~\cite{PX4v114}, while Group~2 further enables DOB-based control~\cite{2024IROS_DOBbased_Yu}.
Each group includes four configurations: \textit{Pre + DOBm}, \textit{Pre only}, \textit{DOBm only}, and \textit{Baseline}, where \textit{Pre} denotes the proposed Pre-Sensing module.
For each configuration, 16 trials are conducted with the object randomly placed approximately 3m away (Fig.~\ref{fig:Exp_C2_Scenario_PassGate} (C)).
By comparing \textit{Pre only} with the baseline and \textit{Pre + DOBm} with \textit{DOBm only}, the ablation design isolates the effects of pre-sensing and DOB.
Overall, the results (Tab.~\ref{tab:ablation_combined}) show that Pre-Sensing consistently improves tracking performance and further enhances performance when combined with DOB-based control, demonstrating its independent and complementary role in the proposed framework.

\begin{table}[h]
\caption{Ablation results on tracking performance}
\label{tab:ablation_combined}
\centering
\renewcommand{\arraystretch}{1.12}
\resizebox{\linewidth}{!}{%
\begin{tabular}{lcccc}
\toprule
\textbf{Method} & \multicolumn{2}{c}{\textbf{Position}} & \multicolumn{2}{c}{\textbf{Attitude}} \\
\cmidrule(lr){2-3}\cmidrule(lr){4-5}
              & \textbf{RMSE [m]} & \textbf{Max [m]} & \textbf{RMSE [rad]} & \textbf{Max [rad]} \\
\midrule
\multicolumn{5}{c}{\textbf{Group 1: PX4 without DOB Control as Baseline}} \\
\midrule
\textbf{Pre + DOBm}    
  & 0.064 (\(\downarrow\)26.4\%) & 0.141 & 0.062 (\(\downarrow\)36.1\%) & 0.298 \\
\textbf{Pre only}      
  & 0.071 (\(\downarrow\)18.4\%) & 0.150 & 0.058 (\(\downarrow\)40.2\%) & 0.293 \\
\textbf{DOBm only / Baseline}$^{\ast}$
~\cite{PX4v114} 
  & 0.087 & 0.291 & 0.097 & 0.307 \\
\midrule
\multicolumn{5}{c}{\textbf{Group 2: PX4 with DOB Control as Baseline}} \\
\midrule
\textbf{Pre + DOBm}    
  & 0.052 (\(\downarrow\)26.8\%) & 0.137 & 0.042 (\(\downarrow\)37.3\%) & 0.193 \\
\textbf{Pre only}      
  & 0.059 (\(\downarrow\)16.9\%) & 0.151 & 0.046 (\(\downarrow\)31.3\%) & 0.170 \\
\textbf{DOBm only / Baseline}$^{\ast}$
~\cite{2024IROS_DOBbased_Yu} 
  & 0.071 & 0.147 & 0.067 & 0.220 \\
\bottomrule
\end{tabular}
}
\vspace{2pt}
{\footnotesize %
$^{\ast}$ Reference values. \(\downarrow\) denotes reduction within each group. \textit{DOBm only / Baseline}$^{\ast}$ indicates identical results for the DOBm-only case and the baseline.}
\vspace{-0.6cm}
\end{table}

\section{CONCLUSION AND FUTURE WORK}
\label{sec:Conclusion_and_Discussion}
This paper presents a unified aerial manipulation system that pre-senses object inertial properties and adaptively adjusts control in real time. The system integrates a vision-language-based human-inspired inertia estimator and an inertia-aware adaptive strategy, achieving $<3\%$ inertia estimation error within 2~s in average and improving position and attitude tracking accuracy by 43\% and 30\% (RMSE), respectively. 

This work paves the way for future advances. Future work will pursue three directions: (i) leveraging the modular design of our perception and control components to deploy the framework on aerial vehicles with varying structures, enabling cross-platform generalization; (ii) refine object prompts online and embed anticipatory force–inertia predictions into the planner so that trajectories pre-compensate for expected disturbances; and (iii) scale the system to multi-object, outdoor settings to test real-time robustness. 

\bibliographystyle{IEEEtran} 
\bibliography{reference} 
\end{document}